\documentclass[runningheads]{llncs}

\usepackage{eccv}

\usepackage{eccvabbrv}

\usepackage{graphicx}
\usepackage{booktabs}
\usepackage{multicol}
\usepackage{multirow}
\usepackage[table]{xcolor}
\usepackage{pifont}

\usepackage[accsupp]{axessibility}  

\usepackage{hyperref}

\usepackage{orcidlink}
\newcommand\blfootnote[1]{%
  \begingroup
  \renewcommand\thefootnote{}%
  \footnotetext{#1}%
  \endgroup
}

\begin{document}

\title{CMCC-ReID: Cross-Modality Clothing-Change Person Re-Identification} 

\titlerunning{CMCC-ReID: Cross-Modality Clothing-Change Person Re-Identification}

\author{Haoxuan Xu\inst{1} \and
 Hanzi Wang\inst{2} \and
 Guanglin Niu\inst{1}\thanks{Corresponding author.}}

\authorrunning{H.~Xu et al.}

\institute{School of Artificial Intelligence, Beihang University, Beijing, China \and
School of Informatics, Xiamen University, Xiamen, Fujian, China\\
\email{\{xhaoxuan,beihangngl\}@buaa.edu.cn}\\
\textcolor{red}{\url{https://github.com/Xuan266/CMCC-ReID}}}

\maketitle
\blfootnote{This work is partially supported by the National Natural Science Foundation of China (No. U25A20531 and No. 62376016).}

\begin{abstract}
  Person Re-Identification (ReID) faces severe challenges from modality discrepancy and clothing variation in long-term surveillance scenario. While existing studies have made significant progress in either Visible-Infrared ReID (VI-ReID) or Clothing-Change ReID (CC-ReID), real-world surveillance system often face both challenges simultaneously. To address this overlooked yet realistic problem, we define a new task, termed Cross-Modality Clothing-Change Re-Identification (CMCC-ReID), which targets pedestrian matching across variations in both modality and clothing. To advance research in this direction, we construct a new benchmark SYSU-CMCC, where each identity is captured in both visible and infrared domains with distinct outfits, reflecting the dual heterogeneity of long-term surveillance.
  To tackle CMCC-ReID, we propose a Progressive Identity Alignment Network (PIA) that progressively mitigates the issues of clothing variation and modality discrepancy. Specifically, a Dual-Branch Disentangling Learning (DBDL) module separates identity-related cues from clothing-related factors to achieve clothing-agnostic representation, and a Bi-Directional Prototype Learning (BPL) module performs intra-modality and inter-modality contrast in the embedding space to bridge the modality gap while further suppressing clothing interference. Extensive experiments on the SYSU-CMCC dataset demonstrate that PIA establishes a strong baseline for this new task and significantly outperforms existing methods.
\end{abstract}

\section{Introduction}
\label{sec:intro}

Person Re-Identification (ReID) \cite{c15,c16,c17,c56,d0} aims to match pedestrian images captured across different non-overlapping cameras, and has become a fundamental task in intelligent surveillance and public security applications. Despite the remarkable progress achieved on controlled benchmarks, traditional ReID methods struggle to generalize well in unconstrained real-world environments, particularly under long-term surveillance conditions.

\begin{figure}[t]
    \centering
    \includegraphics[width=0.6\linewidth]{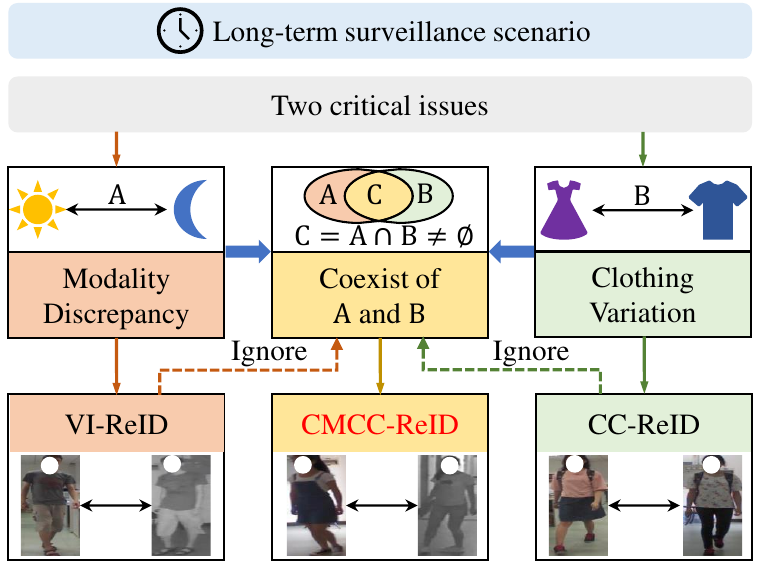}     
\caption{Motivation of CMCC-ReID. In the long-term surveillance scenarios, two critical issues (\textit{i.e.}, modality discrepancy and clothing variation) are not mutually exclusive but coexist. This observation motivates the CMCC-ReID task, which targets pedestrian matching across both modality and clothing changes.}
\label{fig:1}
\end{figure}

In such long-term surveillance scenario, two critical issues jointly challenge conventional ReID systems: active clothing variations resulting from intentional outfit changes and passive modality discrepancies induced by illumination transitions between day and night. As illustrated in Fig.~\ref{fig:1}, existing research typically isolates these challenges into two separate settings: Visible–Infrared Re-ID (VI-ReID) \cite{c18,c19} assumes fixed clothing conditions, while Clothing-Change Re-ID (CC-ReID) \cite{c0,c1} assumes consistent good illumination. However, these assumptions are overly idealized, as they overlook the fact that the two issues are not mutually exclusive and often coexist in real-world scenarios. This observation motivates us to investigate a more realistic and challenging setting, termed \textbf{C}ross-\textbf{M}odality \textbf{C}lothing-\textbf{C}hange \textbf{Re-ID} (\textbf{CMCC-ReID}), which requires jointly addressing both modality discrepancy and clothing variation for reliable long-term identity association.

To advance research on the CMCC-ReID task, we construct a new dataset, termed \textbf{SYSU} \textbf{C}ross-\textbf{M}odality \textbf{C}lothing-\textbf{C}hange (\textbf{SYSU-CMCC}), derived from PRCC \cite{c1} and SYSU-MM01 \cite{c18}. In SYSU-CMCC, each identity is captured under both visible and infrared modalities, ensuring the presence of modality discrepancies. Moreover, to introduce clothing variations, the visible and infrared images of the same identity are collected under different outfits. The detailed description of the SYSU-CMCC dataset is provided in Sec.~\ref{sec:sysu}.

\begin{figure}[t]
    \centering
    \includegraphics[width=0.85\linewidth]{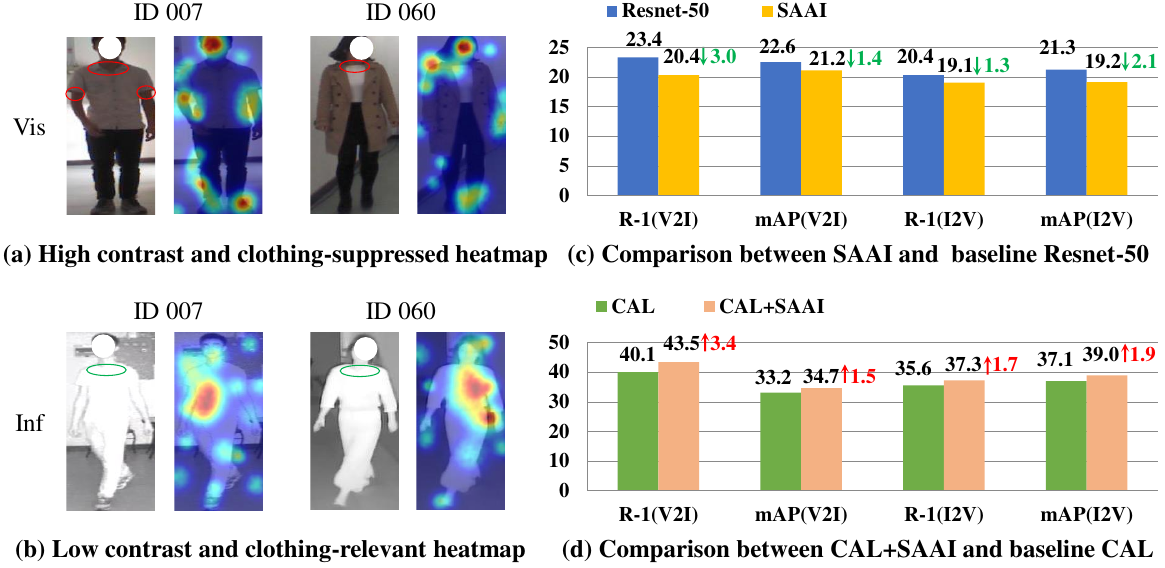}   
\caption{Illustration of the key issues in CMCC-ReID. (a) In the visible modality, the high contrast enables CAL to produce a heatmap with suppressed clothing cues. (b) In the infrared modality, low contrast causes CAL to generate a heatmap dominated by clothing information. (c) Comparison between SAAI and its ResNet-50 baseline shows that direct modality alignment may even degrade performance under clothing variation. (d) Integrating SAAI into CAL yields moderate improvements once clothing interference is mitigated.}
\label{fig:4}
\end{figure}
The core challenge of CMCC-ReID lies in learning identity-consistent representations under the coexistence of modality discrepancy and clothing variation. Unlike a simple combination of CC-ReID and VI-ReID, CMCC-ReID presents a more entangled setting where these two types of heterogeneity mutually reinforce each other. Beyond the inherent clothing variation and modality gap inherited from the two tasks, CMCC-ReID further introduces two key issues that must be addressed:

\noindent\textbf{(1) Extracting clothing-irrelevant features under information-degraded infrared modality.} Existing CC-ReID methods \cite{c2,c4,c11} typically learn clothing-invariant representations under the visible modality with rich color and texture cues. However, in the infrared modality, the absence of such cues causes the body to appear uniformly bright, resulting in low visual contrast between clothing regions and other body parts. This degradation blurs semantic boundaries and hampers the disentanglement of identity cues from clothing information. To verify this, we visualize the attention maps of CAL \cite{c2} under both modalities. As shown in Fig.~\ref{fig:4}(a), CAL effectively suppresses clothing-related responses when high visual contrast is present. In contrast, under the infrared modality, as shown in Fig.~\ref{fig:4}(b), its attention shifts toward clothing-related regions, hindering the extraction of identity-consistent features. Consequently, although existing methods perform well under the visible spectrum, they fail to generalize to the information-degraded infrared modality, leading to limited performance.

\noindent\textbf{(2) Preventing clothing interference before and during cross-modality alignment.}
Existing VI-ReID methods typically eliminate modality discrepancy by aligning global \cite{c21, c22} or part-level \cite{c32, c37} representations within shared embedding space. However, in the CMCC-ReID, clothing acts as a noisy factor, performing direct feature alignment without suppressing clothing-related interference can mislead the model to align clothing-related clues instead of true identity semantics.
To empirically validate this, we compare a state-of-the-art VI-ReID method SAAI \cite{c32} with its baseline (\textit{i.e.}, a standard ResNet-50 \cite{c13} network trained solely with an identity classification loss). As shown in Fig.~\ref{fig:4}(c), SAAI performs even inferior to the baseline, suggesting that naive modality alignment may be counterproductive. Whereas, this does not imply that modality alignment is ineffective. When the core module of SAAI is incorporated into CAL \cite{c2}, the performance exhibits a moderate improvement as shown in Fig.~\ref{fig:4}(d). These observations indicate that while modality alignment alone may strengthen appearance bias, it becomes beneficial when conducted upon clothing-irrelevant representations. Therefore, suppressing clothing interference before and during cross-modality alignment is crucial for CMCC-ReID.

To this end, we propose a novel \textbf{P}rogressive \textbf{I}dentity \textbf{A}lignment Network (\textbf{PIA}) that progressively disentangles and aligns identity representations across both clothing and modality variations. The central idea of PIA lies in a progressive learning paradigm that first constructs clothing-irrelevant identity representations in Stage I and subsequently aligns them across modalities in Stage II. 
In Stage I, a \textbf{D}ual-\textbf{B}ranch \textbf{D}isentanglement \textbf{L}earning (\textbf{DBDL}) module is proposed to explicitly separate identity-related cues from clothing-related noisy through a dual-branch architecture combined with an orthogonality constraint. In Stage II, a \textbf{B}i-Directional \textbf{P}rototype \textbf{L}earning (\textbf{BPL}) module is introduced to jointly perform intra-modality contrast to further refine identity consistency and inter-modality contrast to reduce the modality gap within a prototype learning scheme.
By jointly optimizing the DBDL and BPL modules under the progressive learning strategy, PIA evolves from disentangled feature extraction to modality alignment, achieving consistent and discriminative identity representations.

Our main contributions are summarized as follows:
\begin{itemize}
\item We introduce a new and realistic task CMCC-ReID, which jointly addresses clothing variation and modality discrepancy. To advance this study, we construct the SYSU-CMCC dataset, where each identity is captured under distinct modalities and disjoint clothing conditions.
\item We propose PIA for effective CMCC-ReID, focusing on extracting identity-consistent representations through a progressive learning paradigm.
\item We propose a DBDL module for clothing-irrelevant feature extraction in both modalities and a BPL module for effective cross-modality alignment while further suppressing clothing interference.
\item Extensive experiments validate that PIA establishes a strong baseline and significantly outperforms existing CC-ReID and VI-ReID methods.
\end{itemize}

\section{Related Work}
\label{sec:relatedwork}
\subsection{Visible-Infrared Person ReID} 
VI-ReID~\cite{c18,c19,a5} aims to match pedestrian images across different modalities. To mitigate the modality gap, extensive studies~\cite{c42,c45,c46,c24,c25,c33,c55} have explored learning modality-shared representations through various strategies, including auxiliary modality construction~\cite{c20,c21,c22,c28,c47}, feature alignment~\cite{c26,c31,c34,c44,c38}, modality compensation~\cite{c27,c35,c36,c40,c41,d1}, metric learning~\cite{c30,c29,c43}, and generative translation~\cite{c39,c23,c48,d2}. Moreover, several part-based approaches~\cite{c31,c32,c37} further enhance cross-modality discrimination by enforcing local semantic consistency, thereby reducing intra-class variance.
However, existing VI-ReID methods are developed under the assumption of clothing consistency, overlooking the fact that clothing variation and modality discrepancy often coexist in real-world scenarios, which limits their practical applicability. Moreover, conventional VI-ReID approaches that focus primarily on modality alignment cannot be directly extended to the CMCC-ReID task. Without explicit constraints to suppress clothing interference, these methods tend to overfit to appearance-specific patterns, leading to unstable identity representations across modalities and clothing conditions.

A recent study~\cite{c59} extends VI-ReID to clothing-change scenarios but treats clothing variation as a supplementary factor rather than systematically analyzing its coupling with modality discrepancy. Furthermore, their method relies on lightweight adaptations of existing VI-ReID baselines, whereas our PIA introduces a progressive learning framework incorporating DBDL and BPL to disentangle modality and clothing interference in a stage-wise manner.

\subsection{Clothing-Change Person ReID}
CC-ReID~\cite{c0,c1,c49,c50,c51,a6} aims to retrieve pedestrian images across different outfits. To extract clothing-irrelevant representations, existing methods~\cite{c3,c10,c11,c12,c53, c60} enhance identity discrimination through either multi-modality guidance or single-modality disentanglement. The former exploits auxiliary cues such as sketches~\cite{c1}, keypoints~\cite{c0}, human contours~\cite{c4}, gait patterns~\cite{c5}, or 3D body geometry~\cite{c6,c9,c54} to mitigate clothing interference. The latter focuses on learning clothing-agnostic features within a single modality via adversarial learning~\cite{c2}, status awareness~\cite{c7}, or causal inference~\cite{c8}. 
Beyond the limited applicability similar to VI-ReID, existing CC-ReID methods cannot be directly generalized to the CMCC-ReID task. Due to the information degradation in the infrared modality, these methods struggle to suppress clothing interference, resulting in suboptimal performance under cross-modality and clothing-change conditions.

\section{The SYSU-CMCC Dataset}
\label{sec:sysu}
To advance research on the CMCC-ReID task, we construct a new dataset named SYSU-CMCC, which combines the heterogeneous characteristics of visible and infrared modalities with clothing variations within the same identity. 

\begin{figure*}[t]
    \centering
    \includegraphics[width=0.75\linewidth]{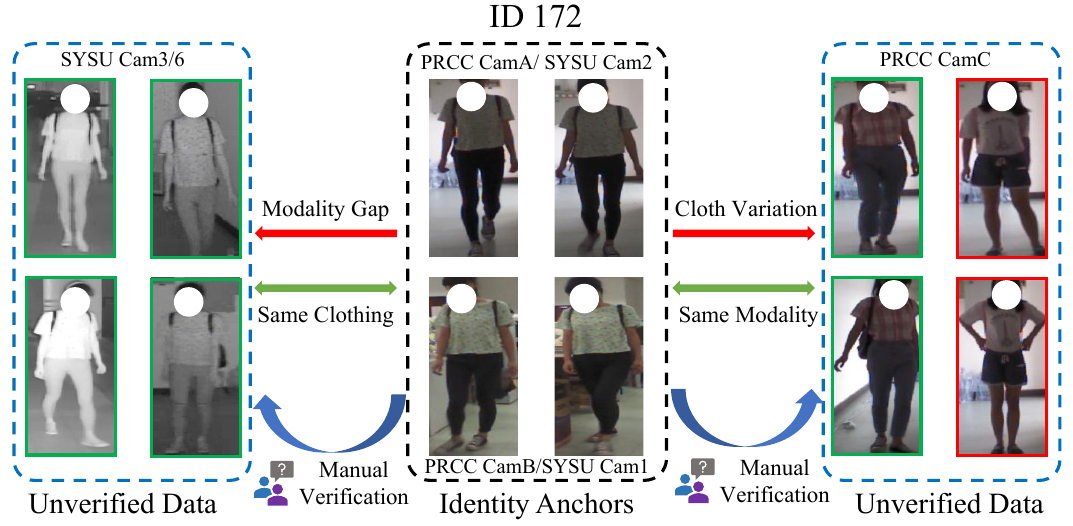}     
\caption{Cross-dataset verification protocol for SYSU-CMCC construction. Samples from PRCC camA/B and SYSU-MM01 cam1/2 serve as \textbf{identity anchors} due to their identical content. For remaining samples (PRCC camC and SYSU-MM01 cam3/6), manual verification is performed: inconsistent samples are marked with \textcolor{red}{red} bounding boxes and discarded, while consistent samples are highlighted with \textcolor{green}{green} bounding boxes and retained to form the SYSU-CMCC dataset.}
\label{fig:verification}
\end{figure*}

\begin{table}[t]
\centering
\small
\caption{Statistical comparisons between the SYSU-CMCC dataset with existing CC-ReID and VI-ReID datasets.}
\renewcommand{\arraystretch}{1}
\setlength{\tabcolsep}{2pt}
\begin{tabular}{c|c|c|c|c|c}
\hline
Dataset&Clothing-change&Cross-modality &IDs& Images & Cameras \\\hline
PRCC~\cite{c1}&\ding{52}&\ding{55}&221&33,698&3\\
LTCC~\cite{c0}&\ding{52}&\ding{55}&152&17,138&12\\
RegDB~\cite{c58}&\ding{55}&\ding{52}&412&8240&2\\
SYSU-MM01~\cite{c18}&\ding{55}&\ding{52}&491&38,271&6\\\hline
SYSU-CMCC&\ding{52}&\ding{52}&214&18,375&3\\\hline
\end{tabular}
\label{tab:statistic}
\end{table}

\noindent\textbf{Dataset Construction.} The SYSU-CMCC dataset is constructed by integrating PRCC and SYSU-MM01, where visible images from PRCC and infrared images from SYSU-MM01 represent two distinct clothing appearances per identity. Since both source datasets inevitably contain \textbf{annotation noise} and \textbf{low-quality samples}, and share overlapping identity labels, we adopt a manual cross-dataset verification protocol to ensure identity consistency. As illustrated in Fig.~\ref{fig:verification}, samples from PRCC camA/B and SYSU-MM01 cam1/2 are identical, serving as reliable identity anchors across the two datasets. Building upon these anchors, we further perform manual verification on samples from PRCC camC and SYSU-MM01 cam3/6. Samples violating identity consistency are marked with red bounding boxes and removed, while those satisfying consistency are highlighted with green bounding boxes and retained to construct the final SYSU-CMCC dataset. This construction ensures that each identity simultaneously exhibits modality discrepancy and clothing variation, while the rigorous filtering process significantly improves annotation quality and dataset reliability.

\noindent\textbf{Dataset Description and Evaluation Protocol.} As summarized in Tab.~\ref{tab:statistic}, SYSU-CMCC comprises 214 identities and 18,375 images, encompassing diverse clothing styles, illumination conditions, and camera viewpoints. Specifically, pedestrians captured by Cam1/2 are under the infrared modality wearing clothing A, while the same identities in Cam3 are imaged under the visible modality with clothing B. During testing, we adopt the Visible to Infrared (V2I) and Infrared to Visible (I2V) matching protocols to evaluate cross-modality retrieval under clothing variation. The SYSU-CMCC dataset provides the first benchmark for jointly studying modality discrepancy and clothing variation, offering a platform for developing CMCC-ReID.

\section{Methodology}
\label{sec:method}
\begin{figure*}[t]
    \centering
    \includegraphics[width=1.0\linewidth]{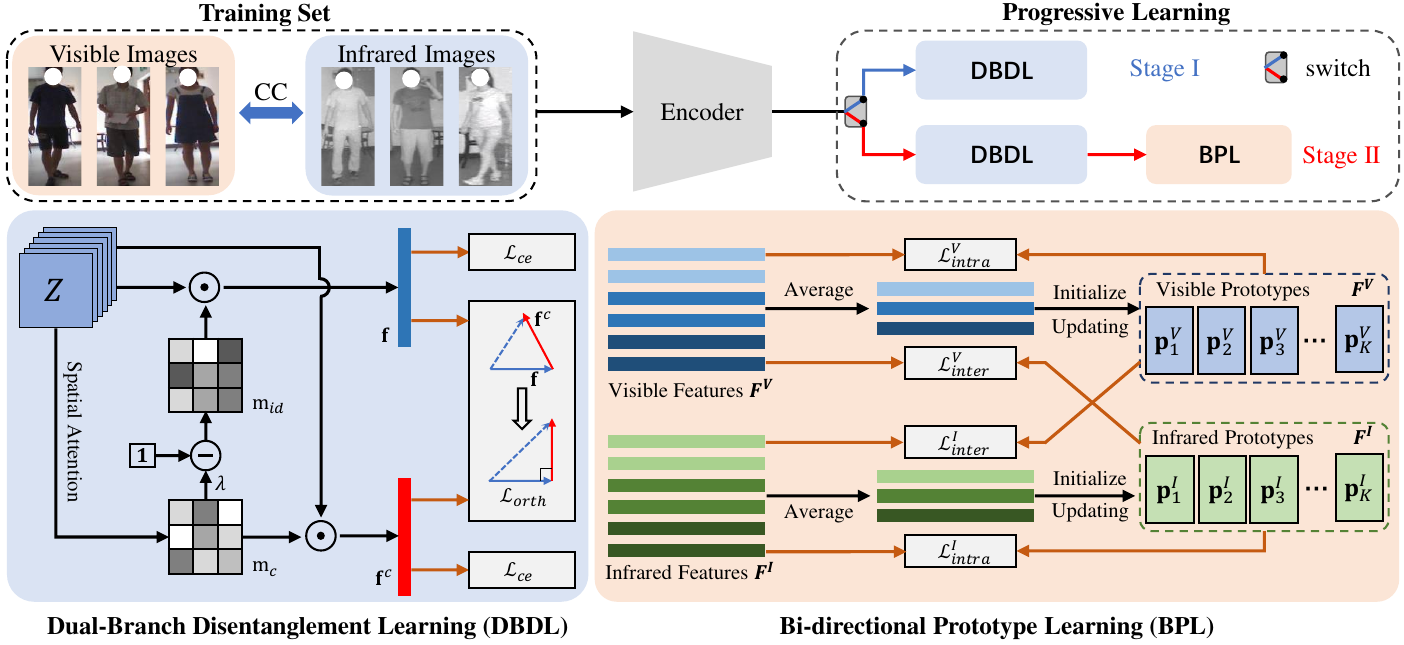}     
\caption{Overview of PIA, which consists of two key components: a Dual-Branch Disentanglement Learning (DBDL) module for clothing-invariant feature extraction in both modalities, and a Bi-Directional Prototype Learning (BPL) module for cross-modality identity alignment. Through progressive optimization, PIA achieves robust and modality-consistent identity representation.}
\label{fig:2}
\end{figure*}

In this section, we present the details of our proposed PIA. As illustrated in Fig.~\ref{fig:2}, PIA is built upon a convolutional backbone. To effectively suppress clothing interference before cross-modality alignment, which is crucial for CMCC-ReID, PIA adopts a progressive learning paradigm that first mitigates clothing variation and then bridges modality discrepancies. Moreover, PIA comprises two key components: a DBDL module that extracts identity-consistent and clothing-agnostic representations, participating in both Stages, and a BPL module that aligns identity embeddings across modalities, which is introduced in Stage II.

\subsection{Dual-Branch Disentanglement Learning}

In CMCC-ReID, both visible and infrared modalities inevitably contain clothing-related cues that interfere with identity recognition. Regardless of the modality, identity and clothing can be regarded as two semantically independent factors: the appearance of clothing should not influence the inherent identity representation, and changes in identity should not affect the representation of the same outfit. Based on this insight, we propose the DBDL module, which explicitly models identity and clothing features through dual-branch feature learning. Furthermore, it enforces orthogonality between the identity and clothing feature subspaces, ensuring that identity features and clothing features do not mutually interfere with each other.

Formally, given a training sample \(x_i\), we denote its identity label as \(y_i^{id}\) and its clothing label as \(y_i^{c}\), where the latter one is defined as a fine-grained class following \cite{c2}.
For an input image \(x_i\), its global feature map \(\mathbf{Z} \in \mathbb{R}^{C \times H \times W}\) extracted by the shared convolutional backbone serves as the input to DBDL. To guide the clothing branch in effectively capturing clothing-related cues, we incorporate a lightweight spatial attention mechanism~\cite{c52} to generate a clothing-related spatial weight mask \(\mathbf{m}_c\), formulated as:
\begin{equation}
    \mathbf{m}_c = \sigma\left(\mathbf{W}_{1} \ast \left[\text{mp}(\mathbf{Z}); \text{ap}(\mathbf{Z})\right]\right),
\end{equation}
where \(\text{mp}(\cdot)\) and \(\text{ap}(\cdot)\) denote max pooling and average pooling along the channel dimension, respectively, \(\ast\) represents the convolution operation, \(\mathbf{W}_{1}\) is the convolution filter weight, and \(\sigma(\cdot)\) denotes the sigmoid activation.

After obtaining the clothing-related spatial weight mask \(\mathbf{m}_c\), the next step is to guide the identity branch to emphasize clothing-irrelevant regions while suppressing clothing-related ones. A straightforward solution would be to directly invert \(\mathbf{m}_c\) to construct the identity attention map. However, such a hard exclusion may inevitably remove implicit identity cues that are still present within clothing regions, thereby weakening the model’s discriminative ability.

Consequently, we introduce a smooth suppression strategy that adaptively balances the influence of clothing areas through a learnable coefficient \(\lambda\). Specifically, we generate the identity-related spatial weight mask \(\mathbf{m}_{id}\) as:
\begin{equation}
    \mathbf{m}_{id} = \mathbf{1} - \lambda \cdot \mathbf{m}_c,
\end{equation}
where $\mathbf{1}$ denotes an all-one matrix with the same dimensions as $\mathbf{m}_c$, $\lambda$ is a learnable parameter that controls the degree of suppression on clothing regions. This design allows the network to retain a small proportion of potentially informative identity details within clothing areas while effectively reducing their interference. 

After obtaining the spatial weight masks $\mathbf{m}_c$ and $\mathbf{m}_{id}$, we can obtain the disentangled identity features $\mathbf{f}$ and clothing-related features $\mathbf{f}^c$, which can be formulated as:
\begin{equation}  
    \mathbf{f} = \text{P}(\mathbf{m}_{id} \odot \mathbf{Z}), \quad
    \mathbf{f}^c = \text{P}(\mathbf{m}_c \odot \mathbf{Z}), 
\end{equation}
where $\odot$ denotes element-wise multiplication, $\text{P}(\cdot)$ denotes the global pooling operation.

Each branch is supervised by an individual classification objective to ensure discriminative representation learning for both identity and clothing. Specifically, cross-entropy loss functions are applied to the identity and clothing features, respectively:
\begin{equation}
    \mathcal{L}_{ce} = -\frac{1}{N} \sum_{i=1}^{N} \log p(y_i^{id}|\mathbf{f}_i; \theta_{id}) 
                      -\frac{1}{N} \sum_{i=1}^{N} \log p(y_i^{c}|\mathbf{f}_i^{c}; \theta_{c}),
\end{equation}
where $N$ denotes the batch size, $p(\cdot|\cdot; \theta)$ denotes the predicted class probability parameterized by the corresponding classifier head weights $\theta_{id}$ and $\theta_{c}$ for identity and clothing branches, respectively. 

To further encourage the semantic disentanglement between identity and clothing representations, we introduce an orthogonal constraint loss that minimizes the correlation between the two feature spaces:
\begin{equation}
    \mathcal{L}_{orth}={\frac{1}{N}}\sum_{i=1}^{N}{\frac{\left| \left<\mathbf{f}_i,\mathbf{f}_i^{c}  \right> \right|}{\left\| \mathbf{f}_i \right\|_2 \cdot \left\| \mathbf{f}_i^{c} \right\|_2}},
\end{equation}
where $\left| \left<\cdot, \cdot \right> \right|$ represents the absolute value after the dot product of two feature embeddings. This constraint enforces the two subspaces to be decorrelated, allowing the network to better capture identity-consistent semantics while suppressing clothing-related interference. 

\subsection{Bi-Directional Prototype Learning}
After obtaining the disentangled identity features from the DBDL module, the next challenge is to achieve cross-modality alignment while further preventing clothing interference. To this end, we propose a BPL module to achieve both intra-modality compactness and inter-modality consistency within the same identity. Specifically, BPL simultaneously aggregates identity features within each modality and aligns each sample with its corresponding prototype in the other modality, effectively reinforcing identity consistency under dual heterogeneity.  

\noindent\textbf{Notation and Preliminaries.} To facilitate the description of our method, the disentangled identity features $\mathbf{f}$ extracted from a training batch are divided into a visible feature set $F^V = \{\mathbf{f}_1^{V}, \ldots, \mathbf{f}_M^{V}\}$ and an infrared feature set $F^I = \{\mathbf{f}_1^{I}, \ldots, \mathbf{f}_M^{I}\}$,
where $\mathbf{f}_i^{V}$ and $\mathbf{f}_i^{I}$ denote the identity representations of the $i$-th visible and infrared instances, respectively, and $M$ is the number of instances in each modality.
During training, a balanced sampling strategy is employed to ensure that each identity appears with $T$ instances per modality in every mini-batch.

Furthermore, we maintain two modality-specific prototype sets for the visible and infrared modalities:
\begin{equation}
P^{V} = \{\mathbf{p}_1^{V}, \ldots, \mathbf{p}_K^{V}\}, \quad 
P^{I} = \{\mathbf{p}_1^{I}, \ldots, \mathbf{p}_K^{I}\},
\end{equation}
where $\mathbf{p}_k^{V}$ and $\mathbf{p}_k^{I}$ denote the prototypes corresponding to the $k$-th identity in the visible and infrared modalities, and $K$ is the total number of identities in the training set.

\noindent\textbf{Modality-specific Prototype Initialization.}
Each prototype is initialized by averaging the identity features of all samples belonging to the same identity within the corresponding modality:
\begin{equation}
    \mathbf{p}_k^{V} = \frac{1}{T} \sum_{y_i^{id}=k} \mathbf{f}_{i}^{V}, \quad \mathbf{p}_k^{I} = \frac{1}{T} \sum_{y_i^{id}=k} \mathbf{f}_{i}^{I},
\end{equation}
where $T$ is the number of IR/RGB instances per identity in a batch, $y_i^{id}$ is the identity label of the $i$-th instance.

\noindent\textbf{Momentum Updating.} During training, the prototypes are dynamically updated each iteration following a momentum-based strategy. 
Given the mini-batch mean feature of identity \(k\) in modality \(m \in \{V,I\}\), denoted as 
\(\bar{\mathbf{f}}_{k}^{m}\) (computed from samples in the current batch with \(y_i^{id}=k\)), 
the update rule is defined as:
\begin{equation}
    \mathbf{p}_k^{V}(\delta) \leftarrow \alpha\,\mathbf{p}_k^{V}(\delta-1) + (1-\alpha)\,\bar{\mathbf{f}}_{k}^{V},
\label{eq8}
\end{equation}
\begin{equation}
    \mathbf{p}_k^{I}(\delta) \leftarrow \alpha\,\mathbf{p}_k^{I}(\delta-1) + (1-\alpha)\,\bar{\mathbf{f}}_{k}^{I},
\label{eq9}
\end{equation}
where \(\alpha\) is the momentum coefficient and \(\delta\) denotes the current training iteration. 
This momentum mechanism stabilizes the prototype evolution by suppressing batch-level noise 
and enables each prototype to gradually approximate the global identity representation, 
leading to smoother convergence and improved cross-modality generalization.

\noindent\textbf{Intra-modality Prototype Learning.}  
Within the same modality, large intra-class variations may still exist due to clothing changes, pose variations, and etc, which can undermine the compactness of identity representations. To alleviate this issue, we encourage each sample to be pulled closer to its corresponding identity prototype while being pushed away from other prototypes within the same modality. 
Formally, the intra-modality prototype loss is formulated in a ProtoNCE \cite{a4} manner as:
\begin{equation}
    \mathcal{L}_{intra}^V = -\frac{1}{M}\sum_{i=1}^{M}\log \frac{\exp\big(\mathbf{f}_i^{V}.\,\mathbf{p}_{y_i^{id}}^{V}/\tau\big)}{\sum_{k=1}^{K}\exp\big(\mathbf{f}_i^{V}.\,\mathbf{p}_{k}^{V}/\tau\big)},
\end{equation}
\begin{equation}
    \mathcal{L}_{intra}^I = -\frac{1}{M}\sum_{i=1}^{M}\log \frac{\exp\big(\mathbf{f}_i^{I}.\,\mathbf{p}_{y_i^{id}}^{I}/\tau\big)}{\sum_{k=1}^{K}\exp\big(\mathbf{f}_i^{I}.\,\mathbf{p}_{k}^{I}/\tau\big)},
\end{equation}
where $\tau$ is a temperature hyperparameter, which is fixed to $1/16$ following~\cite{a3}.

\noindent\textbf{Inter-modality Prototype Learning.}
While intra-modality prototype learning effectively compacts features within each modality, achieving intra-modality consistency, the next goal is to align identity semantics across modalities.

To ensure identity discrimination in cross-modality retrieval, a visible sample should be closer to its corresponding identity prototype in the infrared space than to any other identity prototypes, and vice versa. Motivated by this intuition, we introduce inter-modality prototype learning, which enforces modality-invariant identity alignment. This mechanism reduces the large appearance discrepancy caused by the modality gap and encourages the model to focus on discriminative identity-related clues. Formally, the inter-modality prototype loss is defined as:
\begin{equation}
    \mathcal{L}_{inter}^V = -\frac{1}{M}\sum_{i=1}^{M}\log \frac{\exp\big(\mathbf{f}_i^{V}.\,\mathbf{p}_{y_i^{id}}^{I}/\tau\big)}{\sum_{k=1}^{K}\exp\big(\mathbf{f}_i^{V}.\,\mathbf{p}_{k}^{I}/\tau\big)},
\end{equation}
\begin{equation}
    \mathcal{L}_{inter}^I = -\frac{1}{M}\sum_{i=1}^{M}\log \frac{\exp\big(\mathbf{f}_i^{I}.\,\mathbf{p}_{y_i^{id}}^{V}/\tau\big)}{\sum_{k=1}^{K}\exp\big(\mathbf{f}_i^{I}.\,\mathbf{p}_{k}^{V}/\tau\big)}.
\end{equation}

\subsection{Progressive Learning}
  
\noindent\textbf{Stage I: Optimize with DBDL.}  
The first stage aims to purify identity representations before any cross-modality alignment. Since aligning noisy features may reinforce clothing-induced biases, we optimize only the DBDL module in this stage to extract clothing-invariant and modality-robust identity cues. The training objective is defined as:
\begin{equation}
\mathcal{L}_{\text{I}} = \mathcal{L}_{base} + \mathcal{L}_{ce} + \lambda_1 \mathcal{L}_{orth},
\label{eq:stage1}
\end{equation}
where $\lambda_1$ regulates the strength of the orthogonality constraint for effective feature disentanglement.

\noindent\textbf{Stage II: Jointly optimize with BPL.}  
Once disentangled identity features are obtained, we introduce cross-modality alignment on top of these purified representations to mitigate modality gap without reintroducing clothing interference. In this stage, the DBDL and BPL modules are jointly optimized: BPL further suppresses clothing-induced noise via intra-modality prototype refinement, while simultaneously bridging spectral discrepancy through inter-modality prototype alignment. The training loss in Stage II is:
\begin{equation}
\mathcal{L}_{\text{II}} = \mathcal{L}_{\text{I}} + \mathcal{L}_{intra}^V + \mathcal{L}_{intra}^I + \lambda_2 (\mathcal{L}_{inter}^V + \mathcal{L}_{inter}^I),
\label{eq:stage2}
\end{equation}
where $\lambda_2$ controls the contribution of intra-modality prototype learning.

\section{Experiments}
\begin{table*}[t]
\small
\centering
\renewcommand{\arraystretch}{1}
\renewcommand\tabcolsep{4pt}
\caption{Comparison with state-of-the-art VI-ReID and CC-ReID methods on the SYSU-CMCC dataset. The best and second-best results are highlighted in bold and underlined.}
\begin{tabular}{c|c|cccc|cccc}
\hline
\multirow{3}{*}{Type} & \multirow{3}{*}{Method} & \multicolumn{8}{c}{SYSU-CMCC} \\ \cline{3-10}
 &  & \multicolumn{4}{c|}{Visible to Infrared} & \multicolumn{4}{c}{Infrared to Visible} \\ \cline{3-10}
 &  & R-1 & R-10 & R-20 & mAP & R-1 & R-10 & R-20 & mAP \\
\hline

\multirow{10}{*}{VI-ReID} 
& DDAG \cite{c46}   & 18.3 & 52.1 & 54.7 & 19.6 & 16.8 & 49.7 & 54.6 & 17.2 \\
& CAJ \cite{c45} & 21.5 & 53.4 & 56.9 & 20.7 & 19.8 & 50.6 & 56.2 & 19.6 \\
& MPANet \cite{c37}  & 19.8 & 51.2 & 55.5 & 19.9 & 18.4 & 51.1 & 57.8 & 19.0 \\
& PMT \cite{c47}  & 18.6 & 50.8 & 57.4 & 19.2 & 17.9 & 52.5 & 58.7 & 18.9 \\
& SAAI \cite{c32} & 20.4 & 52.3 & 57.6 & 21.2 & 19.1 & 50.2 & 58.6 & 19.2\\
& MCLNet \cite{c29}  & 22.6 & 56.2 & 60.3 & 23.4 & 21.4 & 56.4 & 59.4 & 21.7 \\
& HOS-Net \cite{c24}  & 23.1 & 57.5 & 59.6 & 24.3 & 21.7 & 57.2 & 61.4 & 22.7 \\
& MMN \cite{c22} & 21.8 & 54.2 & 56.1 & 21.3 & 20.5 & 55.6 & 59.3 & 19.7\\
& IDKL \cite{c42} & 22.4 & 56.0 & 60.3 & 21.7 & 21.3 & 55.7 & 60.8 & 22.3 \\
& DEEN \cite{c19} & 28.3 & 56.9 & 67.7 & 24.8 & 24.7 & 52.4 & 60.5 & 28.5 \\

\hline

\multirow{5}{*}{CC-ReID} 
& GI-ReID \cite{c5} & 24.5 & 53.4 & 62.1 & 22.6 & 21.3 & 49.7 & 57.8 & 23.1 \\
& AIM \cite{c8} & 37.7 & 58.0 & 67.2 & 28.8 & 28.7 & 43.4 & 49.5 & 28.4 \\
& CAL \cite{c2} & 40.1 & 63.2 & 76.2 & 33.2 & 35.6 & \underline{58.2} & 65.3 & 37.1 \\
& CSCI \cite{c60} & \underline{42.9} & 60.8 & 73.1 & 31.9 & 37.6 & 56.7 & 64.8 & 36.9 \\
& CaAug \cite{c10} & 42.6 & \underline{65.6} & \underline{77.4} & \underline{36.6} & \underline{38.2} & 57.4 & \underline{66.4} & \underline{38.2} \\

\hline

\rowcolor{gray!20}
\multirow{1}{*}{Our} 
& PIA & \textbf{57.0} & \textbf{78.6} & \textbf{85.3} & \textbf{46.5} & \textbf{50.4} & \textbf{65.3} & \textbf{69.6} & \textbf{50.3} \\
\hline
\end{tabular}
\label{tab:sys}
\end{table*}

We conduct comprehensive experiments to evaluate the effectiveness of PIA and compare it with state-of-the-art methods on both VI-ReID and CC-ReID tasks. All evaluations are performed on the SYSU-CMCC dataset.
For quantitative assessment, we adopt the Cumulative Matching Characteristic (CMC) curve and the mean Average Precision (mAP) as evaluation metrics. Following standard protocols, we report the results under V2I and I2V modes.

\subsection{Implementation Details}
All experiments are conducted on a single NVIDIA A100 GPU. For our proposed PIA, we adopt ResNet-50 \cite{c13} pre-trained on ImageNet \cite{c14} as the backbone. Following \cite{c16}, we remove the last spatial downsampling operation to enhance feature granularity. In line with \cite{c7}, we apply global average pooling and global max pooling to the backbone’s output feature map, concatenate the pooled features, and normalize the final feature representation using Batch Normalization \cite{a0}. All input images are resized to $384 \times 192$. Data augmentation includes random horizontal flipping, random cropping, and random erasing \cite{a1}. The model is trained via Adam optimizer \cite{a2} for 90 epochs and the Stage II training loss $\mathcal{L}_{\text{II}}$ is used after 55-th epoch. The learning rate is initialized at $3.5 \times 10^{-4}$ and decayed by a factor of 10 every 30 epochs. Each mini-batch contains 64 images, corresponding to 8 identities, where each identity contributes 4 visible and 4 infrared samples to ensure balanced cross-modality training.

\noindent\textbf{Hyperparameters.} The momentum coefficient $\alpha$ in Eq.~\eqref{eq8} and Eq.~\eqref{eq9} is fixed at $0.9$, while the balancing weights $\lambda_1$ and $\lambda_2$ in Eq.~\eqref{eq:stage1} and Eq.~\eqref{eq:stage2} are set to $0.5$ and $1.5$, respectively. All hyperparameters are selected through grid search.

\subsection{Performance Comparisons and Analysis}

\noindent\textbf{Comparison with State-of-the-Art Methods.} We report the performance comparison between PIA and existing state-of-the-art VI-ReID and CC-ReID methods on the SYSU-CMCC dataset. As shown in Tab.~\ref{tab:sys}, PIA significantly surpasses all existing methods under both V2I and I2V settings. Specifically, PIA achieves the best \textbf{57.0\%}/\textbf{46.5\%}, \textbf{50.4\%}/\textbf{50.3\%} Rank-1/mAP accuracy on V2I setting and I2V setting, respectively, exceeding the best-performing VI-ReID methods by a large margin. Compared with the strongest CC-ReID approaches, PIA still exhibits clear advantages, demonstrating its superior ability to handle modality discrepancy and appearance variation simultaneously in CMCC-ReID.  

\noindent\textbf{Comparison between VI-ReID methods and CC-ReID methods.}
Another remarkable observation is that VI-ReID methods perform significantly worse than CC-ReID ones. This result further validates our earlier finding that modality alignment alone is insufficient to maintain identity consistency under CMCC-ReID, highlighting the necessity of the proposed progressive learning strategy, which suppresses clothing interference before cross-modality alignment to ensure stable identity representation.

\begin{table}[t]
\centering
\small
\caption{Ablation studies of each component over the SYSU-CMCC dataset.}
\renewcommand{\arraystretch}{1.15}
\setlength{\tabcolsep}{1.5pt}
\begin{tabular}{c|c|c|c|c|c|c|cc|cc}
\hline
\multirow{2}{*}{Index} & \multirow{2}{*}{Baseline} & \multirow{2}{*}{DBDL} & \multirow{2}{*}{$\mathcal{L}_{orth}$} & \multirow{2}{*}{Progressive} & \multicolumn{2}{c|}{BPL}  & \multicolumn{2}{c|}{V2I} & \multicolumn{2}{c}{I2V} \\ \cline{6-11}
                       &                       &                    &  &                                    &      $\mathcal{L}_{intra}$      &          $\mathcal{L}_{inter}$       & R-1 & mAP &R-1&mAP \\ \hline
1 & \ding{52} & \ding{55} & \ding{55} &\ding{55} &\ding{55} &\ding{55} & 40.1 & 33.2 &35.6& 37.1 \\
2 & \ding{52} & \ding{52} & \ding{55} &\ding{55} &\ding{55}& \ding{55}& 43.2 & 35.7 &38.6& 38.7 \\
3 & \ding{52} & \ding{52} & \ding{52} &\ding{55} &\ding{55}&\ding{55} & 45.6 & 37.6& 40.5& 39.1\\
4 & \ding{52} & \ding{52} & \ding{52} &\ding{52} &\ding{52}& \ding{55}& 46.9 & 39.5&42.7&41.6 \\
5 & \ding{52} & \ding{52} & \ding{52} & \ding{52}&\ding{52}& \ding{52}& \textbf{57.0} & \textbf{46.5} &\textbf{50.4}&\textbf{50.3}\\ 
6 & \ding{52} & \ding{52} & \ding{52} & \ding{55}&\ding{52}& \ding{52}& 42.8 & 35.4 &42.6&40.7\\\hline
\end{tabular}
\label{tab:abl}
\end{table}

\subsection{Ablation Studies}
We conduct comprehensive ablation studies on the SYSU-CMCC dataset to evaluate the effectiveness of each core component in PIA, including the DBDL module, the orthogonal constraint loss $\mathcal{L}_{orth}$, the intra-modality prototype loss $\mathcal{L}_{intra}$ and inter-modality prototype loss $\mathcal{L}_{inter}$ within the BPL module. A reproduced CAL serves as the baseline, and the detailed results are reported in Tab.~\ref{tab:abl}. Note that all results involving BPL are obtained under the proposed progressive learning scheme.

\begin{figure}[t]
\centering
    \begin{subfigure}{0.48\linewidth}
      \centering 
       \includegraphics[width=\linewidth]{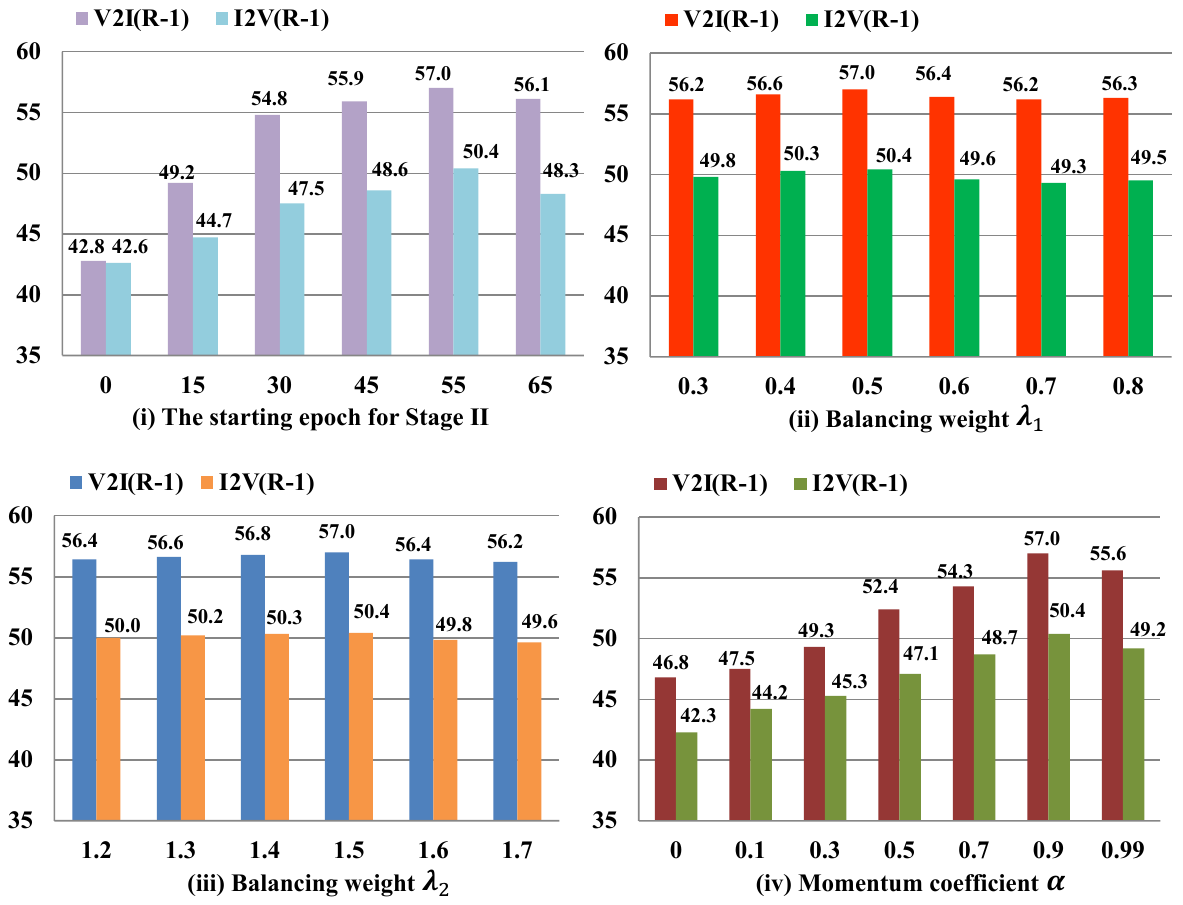}
    
       \caption{The impact of (i) the starting epoch for Stage II, (ii) balancing weight $\lambda_1$ in Eq. (\ref{eq:stage1}), (iii) $\lambda_2$ in Eq. (\ref{eq:stage2}), and (iv) momentum coefficient $\alpha$ in Eq. (\ref{eq8}).}
       \label{fig3}
    \end{subfigure}
    \hfill
    \begin{subfigure}{0.48\linewidth}
      \centering
       \includegraphics[width=\linewidth]{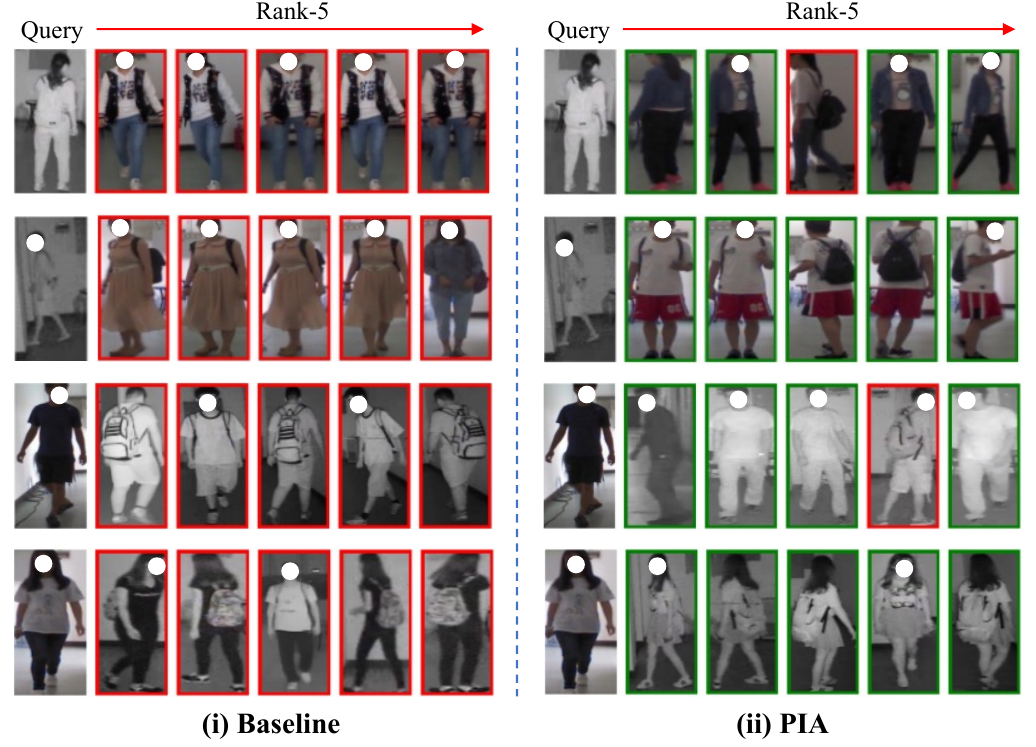}
       \caption{Comparison of Rank-5 retrieval results. The top two rows correspond to the I2V setting, while the bottom two rows correspond to the V2I setting.}
       \label{fig5}
    \end{subfigure}
\caption{(a) Results of parameter sensitivity analysis for key hyperparameters in our model. (b) Visualization of retrieval results.}
\end{figure}

\noindent\textbf{Effectiveness of DBDL.} From Index-1 to Index-2, introducing the DBDL module brings $3.1\%/2.5\%$ and $3.0\%/1.6\%$ improvement in Rank-1/mAP under the V2I and I2V settings, respectively. This demonstrates that DBDL effectively preserves identity-consistent semantics while mitigating clothing interference by modeling two disentangled feature streams. 

\noindent\textbf{Effectiveness of $\mathcal{L}_{orth}$.} Comparing Index-2 and Index-3, adding the orthogonality constraint $\mathcal{L}_{orth}$ further boosts performance. This constraint explicitly enforces the independence between disentangled subspaces, preventing clothing-related cues from leaking into the identity branch. 

\noindent\textbf{Effectiveness of $\mathcal{L}_{intra}$.}
Progressing from Index-3 to Index-4, incorporating $\mathcal{L}_{intra}$ consistently enhances the performance. By compacting identity features within each modality, this loss mitigates intra-modality discrepancies caused by pose, illumination, and clothing variations,resulting in more robust and coherent identity clusters.

\noindent\textbf{Effectiveness of $\mathcal{L}_{inter}$.}
From Index-4 to Index-5, adding $\mathcal{L}_{inter}$ notably improves Rank-1/mAP by $10.1\%/7.0\%$ and $7.7\%/8.7\%$ under the V2I and I2V settings, respectively. This indicates that $\mathcal{L}_{inter}$ plays a critical role in bridging the cross-modality gap. By enforcing modality-invariant identity alignment, it establishes consistent identity semantics. 

\noindent\textbf{Effectiveness of progressive learning.} Comparing Index-6 with Index-3, introducing the BPL without the progressive learning scheme leads to only marginal gains or even performance degradation on certain metrics, highlighting that the effectiveness of PIA primarily stems from the synergistic integration of DBDL, BPL, and the progressive learning framework, all of which are indispensable. To further investigate the impact of the progressive learning paradigm, we vary the epoch at which Stage II training begins and evaluate the corresponding performance, as shown in Fig.~\ref{fig3}(i). The case with 0 epochs corresponds to training without the progressive learning scheme, yielding the lowest accuracy and indicating poor identity-consistent representation under CMCC-ReID. With progressive learning, performance gradually improves as Stage II starts later, achieving the best results when initialized at the 55th epoch, after which performance begins to decline. These findings confirm the effectiveness of the progressive learning strategy.

\noindent\textbf{Hyperparameter Analysis.}
We further conduct a detailed analysis of the hyperparameters under both the V2I and I2V settings on the SYSU-CMCC dataset. As illustrated in Fig.~\ref{fig3}, the model achieves the optimal performance when $\lambda_1$ is set to $0.5$ and $\lambda_2$ to $1.5$. Notably, the overall performance remains relatively stable across a wide range of values, indicating that our PIA is not sensitive to the precise choice of these training balancing weight. 

As for the impact of momentum coefficient $\alpha$ in Eq. (\ref{eq8}), an excessively small or large value can hinder the stability of prototype representations, leading to notable performance degradation. As illustrated in Fig.~\ref{fig3}(iv), the best results are achieved when $\alpha$ is set to 0.9, indicating that a moderate momentum facilitates more stable and discriminative prototype updates.

\begin{figure}[t]
\centering
    \begin{subfigure}{0.48\linewidth}
    \includegraphics[width=\linewidth]{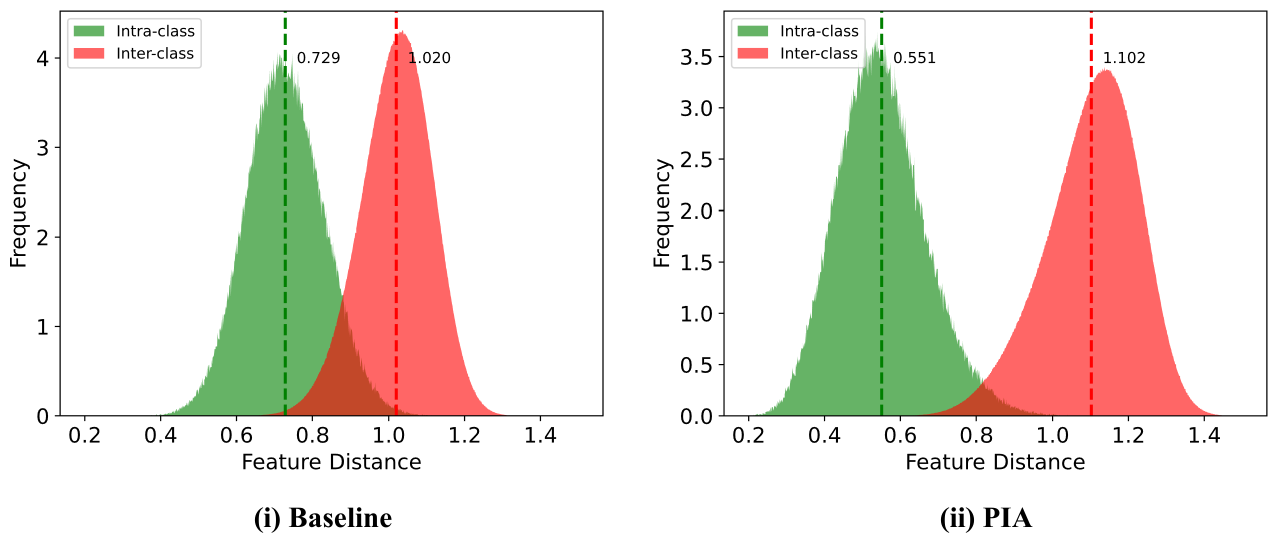}
       \caption{The cosine distance distributions of positive and negative pairs on the SYSU-CMCC dataset during inference.}
       \label{fig6}
    \end{subfigure}
    \hfill
    \begin{subfigure}{0.48\linewidth}
    \includegraphics[width=\linewidth]{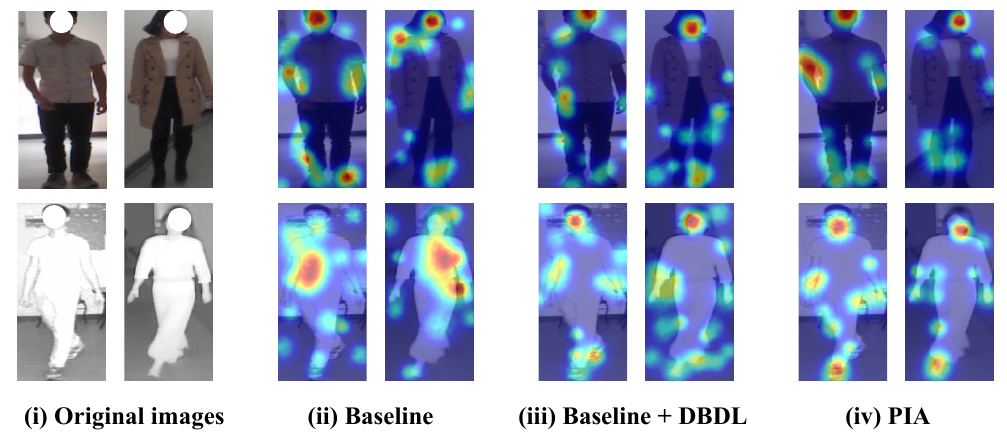}
       \caption{The visualization of attention maps on SYSU-CMCC. (i) original images, (ii) attention maps generated by the baseline CAL, (iii) attention maps from the baseline with DBDL, and (iv) attention maps from the complete PIA.}
       \label{fig7}
    \end{subfigure}
    \caption{Visualizations of (a) cosine distance distributions and (b) attention maps.}
\end{figure}
\subsection{Visualization}
\noindent\textbf{Retrieval Results.} To qualitatively validate the effectiveness of PIA, Fig.~\ref{fig5} presents Rank-5 retrieval comparisons between PIA and the baseline CAL on SYSU-CMCC. For each query, correctly retrieved images are marked with green boxes, while incorrect ones are marked in red. Compared with the baseline, PIA consistently returns more correct matches within the top positions, demonstrating its stronger ability to capture identity cues in CMCC-ReID.

\noindent\textbf{Feature Distribution Analysis.} Fig.~\ref{fig6} shows the cosine distance distributions of positive and negative pairs on SYSU-CMCC during inference. Compared with the baseline CAL, FID notably decreases the mean distance of positive pairs from 0.729 to 0.551, while increasing that of negative pairs from 1.02 to 1.102. 

\noindent\textbf{Attention Maps.} We visualize the attention maps of the baseline model, the baseline with DBDL, and the complete PIA in Fig.~\ref{fig7}. The baseline model exhibits noisy attention introduced by background clutter, failing to suppress clothing-related regions under the infrared modality. With the introduction of DBDL, the model effectively reduces attention to clothing areas and shows improved robustness in information-degraded conditions. Nevertheless, some irrelevant activations still persist in background regions. In contrast, PIA further suppresses both clothing-related and background distractions, producing more concentrated and identity-consistent attention across modalities.

\section{Conclusion}
In this paper, we introduce a new task, termed CMCC-ReID, which aims to match pedestrian images across variations in both modality and clothing, and we establish the SYSU-CMCC dataset, providing the first benchmark for CMCC-ReID. To address CMCC-ReID, we propose a Progressive Identity Alignment Network (PIA) that first disentangles clothing-irrelevant identity features through a novel Dual-Branch Disentanglement Learning (DBDL) module, and then progressively aligns modality representations via the proposed Bi-Directional Prototype Learning (BPL) module. Extensive experiments demonstrate that PIA significantly outperforms existing methods, setting a strong baseline for this challenging task.

\bibliographystyle{splncs04}
\bibliography{main}
\end{document}